\documentclass[review]{elsarticle}
\usepackage[english]{babel}
\usepackage{booktabs}
\usepackage{multirow}
\usepackage{lineno,hyperref}
\usepackage{amsfonts}
\usepackage{dsfont}
\modulolinenumbers[5]
\usepackage[utf8]{inputenc} 
\usepackage{mathtools}
\usepackage{amssymb}
\usepackage[dvipsnames]{xcolor}
\usepackage{lscape}
 \usepackage{tabu}
\usepackage{verbatim}
\usepackage{algorithm}
\usepackage{algorithmic}
\usepackage{calrsfs}
\DeclareMathAlphabet{\pazocal}{OMS}{zplm}{m}{n}

\usepackage{setspace}
\usepackage{multirow}
\usepackage{makecell}
\usepackage{algorithmic}
\usepackage{algorithm}
\usepackage{array, caption, tabularx,  ragged2e,  booktabs}
\journal{Journal of \LaTeX\ Templates}

\journal{Information Sciences}

\begin{document}

\begin{frontmatter}

\title{Automatic Online Multi-Source Domain Adaptation}

\tnotetext[mytitlenote]{Corresponding Author}
\tnotetext[mytitlenote1]{Equal Contribution}

\author[mymainaddress]{Xie Renchunzi$^{**}$}

\author[mymainaddress]{Mahardhika Pratama$^{***}$}


\address[mymainaddress]{School of Computer Science and Engineering, Nanyang Technological University, Singapore}

\begin{abstract}
Knowledge transfer across several streaming processes remain challenging problem not only because of different distributions of each stream but also because of rapidly changing and never-ending environments of data streams. Albeit growing research achievements in this area, most of existing works are developed for a single source domain which limits its resilience to exploit multi-source domains being beneficial to recover from concept drifts quickly and to avoid the negative transfer problem. An online domain adaptation technique under multi-source streaming processes, namely automatic online multi-source domain adaptation (AOMSDA), is proposed in this paper. The online domain adaptation strategy of AOMSDA is formulated under a coupled generative and discriminative approach of denoising autoencoder (DAE) where the central moment discrepancy (CMD)-based regularizer is integrated to handle the existence of multi-source domains thereby taking advantage of complementary information sources. The asynchronous concept drifts taking place at different time periods are addressed by a self-organizing structure and a node re-weighting strategy. Our numerical study demonstrates that AOMSDA is capable of outperforming its counterparts in 5 of 8 study cases while the ablation study depicts the advantage of each learning component. In addition, AOMSDA is general for any number of source streams. The source code of AOMSDA is shared publicly in  \url{https://github.com/Renchunzi-Xie/AOMSDA.git}.     
\end{abstract}


\end{frontmatter}


\section{Introduction}
Multistream classification problem is a research area studying knowledge transfer across many streaming processes \cite{MSC}. It is seen as an extension of conventional transfer learning problem \cite{pan2010survey} where knowledge transfer approach is undertaken from continuously sampled data points calling for special treatment. In addition to the covariate shift problem, the multistream classification problem is complicated by rapid information flow having to be handled with low memory footprint and changing environments happening independently in each stream. This problem exists in daily scenario where data samples are continuously captured in real time. In realm of machine health monitoring problem, data samples stream continuously from sensors. Although data collection is a trivial issue, the labelling process solicits constant operator attention being quite demanding and difficult because it often requires visual inspection leading to frequent stoppages of a manufacturing process. This problem becomes even more problematic than that in the complex manufacturing process involving a number of machines because of possible repetitions of a model building phase across these machines. A plausible solution is to deploy a multistream solution where a model is flexibly transferred across different machines while possessing online and adaptive working principles. 

Several research efforts have been devoted to resolve the issue of multistream classification using a combination of online domain adaptation methods and drift handling techniques \cite{MSC,haque2017fusion,pratama2019atl}. Most of which are crafted for a single source domain setting where its performance depends on the quality of a single information source. Multi-source domains configuration is capable of attracting advantages in dealing with a concept drift where model's performance can be quickly recovered as well as in avoiding the issue of negative transfer \cite{du2019multi}. Model's development for the multi-source domains problem is challenging to fully exploit complementary information of each source domain because it features \textbf{the problem of varying relevance}. That is, relevance of each source domain to target domain is dynamic in nature. This issue worsens in the streaming environments because concept drifts might change the relevance of each source domain. A model is supposed to be selective for an irrelevant source domain while maximizing relevant information of different source domains.    

Several works have been proposed to address the multi-source domains problems \cite{wang2019transferable,zhao2018adversarial,wang2020continuously} but they are not compatible in the streaming environments. To the best of our knowledge, \cite{du2019multi} is the only work in the literature addressing the multi-source domains issue in the streaming environments. Nonetheless, this approach incurs considerable computational and memory burdens because it is based on an ensemble approach. That is, it creates an ensemble classifier for both source and target domains. A new ensemble classifier is created if a drift is detected. Furthermore, this work does not possess a specific domain adaptation strategy thus suffering from limited accuracy in the case of high discrepancy between source and target domains. In a nutshell, multistream classification problem under multi-source streams feature four major issues which have to be tackled simultaneously: 1) \textbf{covariate shift}, which refers to different data distributions of each source stream as well as target stream; 2) \textbf{lack of labelled samples} which happens because labels are only available for source streams while being absent for target stream; 3) \textbf{asynchronous drift} which can be found because concept drifts are independent and take place at different time periods; 4) \textbf{varying relevance} which exists due to changing relationship of source streams to target stream.

An automatic multi-source domain adaptation (AOMSDA) algorithm is proposed in this paper to settle the multistream classification problem under multi-source streams. AOMSDA is designed using the framework of denoising autoencoder (DAE) where the domain adaptation step for the covariate shift issue is developed using shared parameters adjusted in the generative phase minimizing the reconstruction error and in the discriminative phase minimizing the classification error \cite{Zhou2012OnlineIF,ASHFAHANI2020297,pratama2019atl}. The unsupervised domain adaptation is performed here where there does not exist any labelled samples of the target domain while relying solely on labelled samples of source domains. The central moment discrepancy (CMD)-based regularizer is put forward to address the problem of varying relevance in the smooth manner where the domain's discrepancy is measured in the embedding space, transformed space. That is, it determines the confidence degree of a source stream where an irrelevant stream is ignored while accepting those of relevant ones. AOMSDA features a self-organizing structure coping with the asynchronous drift problem. That is, its hidden nodes are grown and pruned in respect to varying distributions of source streams. The node reweighting strategy based on the smoothness concept is applied to address the concept drift in the target domain. 

This paper puts forward four major contributions: 1) this paper resolves multistream classification problem under multi-source streams via algorithmic development of AOMSDA; 2) this paper offers the notion of CMD-based regularizer to handle the varying relevance problem where it sets the confidence degree of each source stream based on its closeness degree to the target stream. It is capable of mixing the complementary information of multi-source domains rather than  only  a  single  source  domain; 3) the concept of node re-weighting strategy is integrated to handle concept drifts of the target stream; 4) the source code of AOMSDA is shared publicly in \url{https://github.com/Renchunzi-Xie/AOMSDA.git} to allow convenient reproduction of our numerical results and further study. AOMSDA's performance has been numerically validated via numerical study in eight problems and comparisons with recently published algorithms. AOMSDA is capable of outperforming other algorithms in five of eight problems with noticeable margin. Furthermore, the advantage of AOMSDA is confirmed further with ablation study and analysis of the number of source domains where each learning component contributes positively to the overall performance of AOMSDA and it is general for any number of source streams. 

The remainder of this paper is structured as follows: Section II discusses related works, Section III introduces the problem setting, Section IV outlines the details of our method, Section V presents our numerical study, and Section VI concludes our paper.

\section{Related Works}
\noindent\textbf{Single Source Domain Adaptation:} the area of unsupervised domain adaptation (UDA) has been an active research topic where it assumes the label availability only in the source domain while leaving the target domain unlabelled. The goal of domain adaptation is to address the issue of covariate shift where there exists a gap between source and target distributions \cite{pan2010survey}. The common approach of domain adaptation makes use of domain discrepancy measure minimized to generate a common feature space of the source and target domains. \cite{TCA}
utilizes the maximum mean discrepancy approach, \cite{zhuang2015supervised} utilizes the Kullback-Leibler (KL) divergence approach and \cite{zellinger2017central} puts forward the central moment discrepancy (CMD) taking into account high order moments. Another approach lies in the adversarial training scenario to establish a domain-invariant representation \cite{ganin2016domain}. It involves the use of a domain classifier classifying the origin of data samples, source or taget while a feature generator aims to fool the domain classifier. The idea of multistream classification aims to enhance the domain adaptation technique in handling streaming data where the covariate shift and the asynchronous drift are handled simultaneously. MSC \cite{MSC} is a pioneering work in this area where it is driven by the kernel mean matching (KMM) approach. A concept drift detector is integrated for each source and target domain where a new classifier is added if a drift is signalled. MSC imposes considerable computational complexity. To correct this shortcoming, FUSION \cite{haque2017fusion} is proposed where it utilizes the KLIEP technique for domain adaptation while the asynchronous drift is alarmed by a density ratio between source and target domain. A deep learning solution of multistream classification problem, namely ATL, is proposed in \cite{pratama2019atl}. A domain-invariant network is attained by the parameter sharing strategy in the generative and discriminative phases of autoencoder (AE) and the KL divergence approach. AOMSDA differs from ATL in the multi-source domains facet where ATL is designed only for the single domain scenario. Furthermore, AOMSDA is equipped by the CMD-based regularizer to address the issue of varying relevance and the node re-weighting strategy to cope with the concept drifts of the target stream. 

Unlike a single source case, the multi-source domain adaptation takes advantage the existence of several source domains which improves the generalization power and prevents the negative transfer problem. Such approach has been proven to be effective in recovering from concept drifts quickly \cite{du2019multi}. Nonetheless, handling several source domains are not easy because of changing relationship between each source domain and the target domain. In addition, each source domain should be combined properly because it might convey complementary, mutually exclusive or even included information. This problem is even more challenging in the streaming environment than in the static case because the asynchronous drift problem might alter the source-to-target relationship.

\noindent\textbf{Multi-Source Domains Adaptation:} multi-source domains adaptation benefits from the existence of multi-source domains allowing it to recover from the concept drift quickly and to avoid the negative transfer problem \cite{du2019multi}. Each source domain might convey complementary, mutually exclusive or included meaning that the underlying challenge lies on how to select or to combine information sources maximizing the performance of target domain suffering from the absence of any labelled samples. One approach is to utilize the weighting strategy \cite{wang2019transferable} where every source domain is weighted in accordance with its relevance to source domain. Another approach is via the adversarial training approach \cite{zhao2018adversarial,wang2020continuously}. As with the single domain setting, a domain classifier is deployed to identify the source of information under a multi-class classification problem. The gradient reversal strategy is implemented along with the normal gradient strategy thus guiding the feature generator to induce a common feature distribution of the target and source domains. \cite{du2019multi} offers a solution of multi-source domains in the streaming context using the idea of ensemble classifiers. It generates a pool of classifiers for every domain and a new pool of classifiers is added if a drift is detected. However, this  approach  suffers from  high  computational  and memory  demands  because  of  the  use  of  ensemble  classifiers for different subsets of every domain. Moreover, this approach is not equipped by a specific domain adaptation strategy thus suffering from performance degradation in the case of high discrepancy between source and target domains.
\section{Problem Formulation}
\subsection{Problem Definition}
\noindent\textbf{Label scarcity Problem}: Label scarcity exists in the multistream classification problem \cite{MSC} because labelled samples are only offered to the source stream while leaving the target stream completely unlabelled. Hence, the underlying challenge is to arrive at decent accuracy in the target stream where no labelled samples exist. This requires an unsupervised domain adaptation strategy taking into account the absence of any labelled samples in the target domain.

\noindent\textbf{Covariate shift problem}: The issue of covariate shift occurs because source and target domains follow different distributions $P_S(X)\neq P_T(X)$. Here, we still assume that source and target domains share the same feature space $X_S,X_T\in\Re^u$ but have different distributions. This issue calls for a domain adaptation strategy such that different distributions can be aligned.

\noindent\textbf{Asynchronous drift problem}: The asynchronous drift problem exists in the multistream classification problem as the nature of data stream. Unlike a single stream case, both source stream and target stream are subject to an independent concept drift occurring at different time periods. A concept drift is defined as the change of conditional distributions where the point of change in the source and target domain is unique: $P_{S}(X,Y)_{t_1}\neq P_{S}(X,Y)_{t_1+1}$ and $P_{T}(X,Y)_{t_2}\neq P_{T}(X,Y)_{t_2+1}$ where $t_1\neq t_2$. This issue requires drift handling mechanism in both source and target domains while still retaining the domain-invariant property. Note that the target domain has no labelled samples.

\noindent\textbf{Varying Relevance Problem}: The varying relevance problem happens because of the dynamic relationship between each source domain and target domain. That is, each source domain might contain complementary, mutually-exclusive or included information. This issue is complicated by the asynchronous drift problem which alters the relationship of each source and target domain.

\subsection{Simulation Procedure}
Multistream classification problem under multi-source domains concerns on a classification problem across many streaming processes running in parallel and independently. All of them except one serve as source streams $B_1^{S_i},B_2^{S_i},...,B_{K_{S_i}}^{S_i}$ while one stream is designated as a target stream $B_1^{T},B_2^{T},...,B_{K_T}^{T}$. $S_i$ stands for the $i-th$ source stream and there are in total $N_s$ source streams of interest. $K_{S_i},K_T$ denote the number of data batches seen thus far $K_S = K_T$. Only source streams are labelled $B_{k_s}^{S_i}=\{x_n,y_n\}_{n=1}^{N_i}$ while the target stream suffer from the absence of any labelled samples $B_{k_t}^T=\{x_n\}_{n=1}^{N_t}$, \textbf{the label scarcity problem}. $N_i,N_t$ respectively denote the size of source and target streams. $x_n\in\Re^{u}$ is an input vector while $y_n\in\{l_1,l_2,...,l_m\}$ is a target vector. $u,m$ respectively label the number of input and output dimensions. The source streams and target stream share the same feature space but feature different distributions, i.e., $X_{S_i},X_{S_j},X_{T}\in\Re^{u}, P_{S_i}(X) \neq P_{S_j}(X) \neq P_{T}(X), i\neq j$ known as \textbf{the covariate shift problem}. Data streams are generated in non-stationary environments leading to \textbf{the asynchronous drift problem}. That is, concept drifts occur at different time indexes $P_{S_i}(X,Y)_{t_1}\neq P_{S_i}(X,Y)_{t_{1}+1},P_{S_j}(X,Y)_{t_2}\neq P_{S_j}(X,Y)_{t_{2}+1},P_{T}(X,Y)_{t_3}\neq P_{T}(X,Y)_{t_{3}+1},t_1\neq t_2\neq t_3$. There exists two types of drifts: virtual drift and real drift \cite{GamaDataStream,KhamassiMouchawehHammamiGhedira16}. The virtual drift refers to changes of the marginal distribution while the real drift pinpoints changes of the conditional distribution. Non-stationary environments also affect the relationship of a source stream and a target stream, i.e., \textbf{the varying relevance problem}. The connection of source streams and a target stream are not constant. 
 \begin{figure*}[ht]
 \centering
 \includegraphics[scale=0.35]{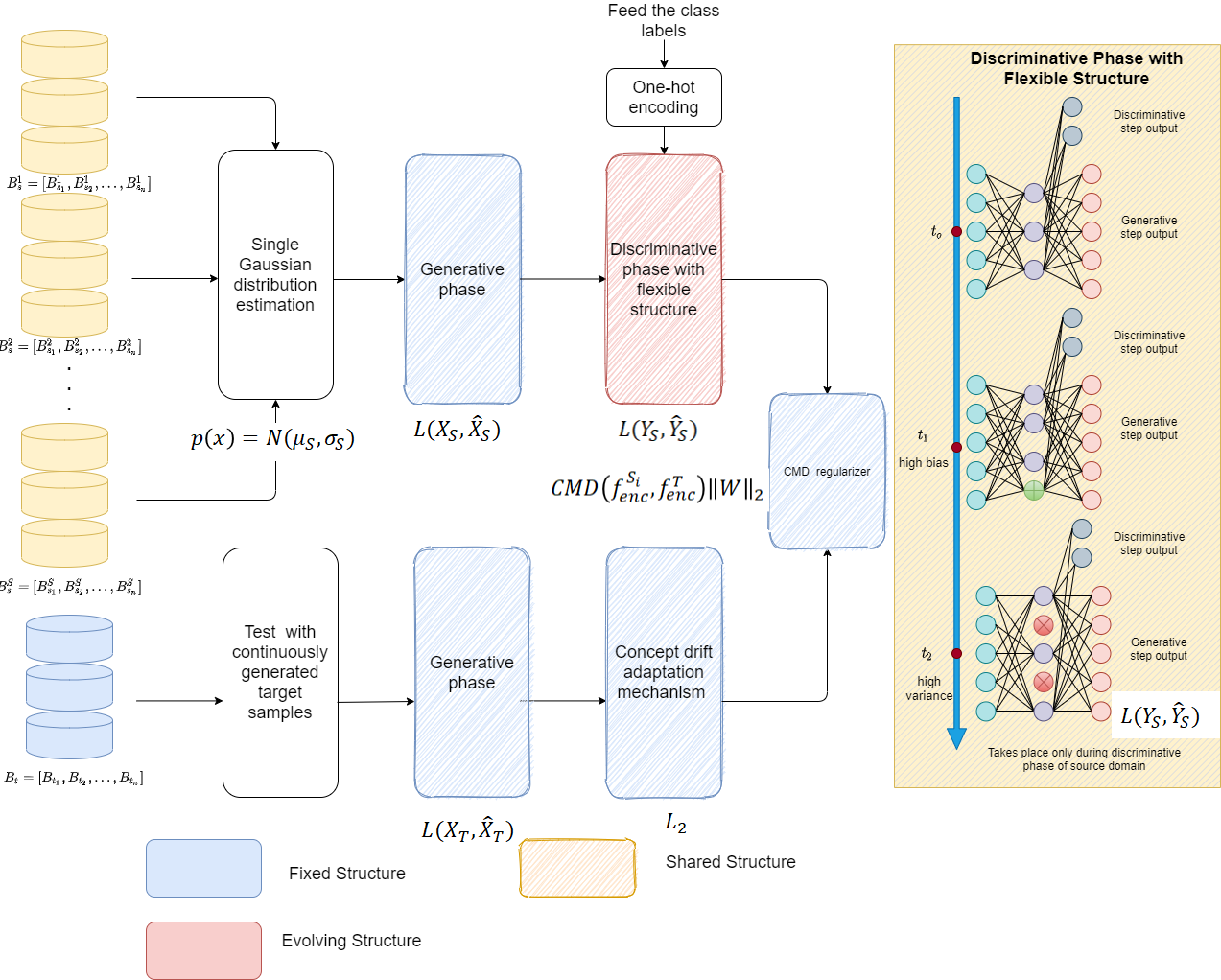}
 \caption{AOMSDA Learning Policy and Network Evolution.}
 \label{fig_framework}
 \end{figure*}
\begin{algorithm}
\caption{AOMSDA's Learning Policy}
\begin{algorithmic}\label{Algorithm_1}
\STATE\textbf{Input}: Source data streams $B_S=[B_1^{S_j},B_2^{S_j},...,B_{N}^{S_j}]$, target data stream $B_T=[B_1^{T},B_2^{T},...B_{N}^{T}]$,  probability density function $p_s=N(\mu_S,\sigma_S)$ , initial network parameters $W=[W_{enc},b_{enc},W_{dec},b_{dec},W_{out},b_{out}]$
\STATE\textbf{Output}: Predicted labels of target data stream $Y_t$.
\FOR{$i=1$ to $N$}
\STATE Update (Source): $p_s=N(\mu_S,\sigma_S)$
\FOR{$j=1$ to $N_s$}
\STATE Test: Predict both $B_{i}^{S_j}, B_{i}^{T}$
\STATE Train (Source): Generative Parameter Learning based on $L(\hat{X}_{S_j}, X_{S_j})$
\STATE Structural Evolution (Source): Structural Learning Mechanism in the discriminative phase using $p_s$
\STATE Train (Source): Discriminative Parameter Learning based on $L(\hat{Y}_{S_j}, Y_{S_j})$
\ENDFOR
\STATE Train (Target): Generative Parameter Learning based on $L(\hat{X}_T, X_T)$
\STATE Train (Target): Node Re-weighting based on $L_2$ (\ref{L2})
\FOR{$j=1$ to $N_s$}
\STATE Train (Source)\&(Target): CMD-based Regularization based on $CMD(f_{enc}^{S_j},f_{enc}^T)$
\ENDFOR
\ENDFOR
\end{algorithmic}
\end{algorithm}

\section{Learning Policy of AOMSDA}
AOMSDA is developed to handle the multistream classification problem under multi-source domains condition having four bottlenecks: the covariate shift, the asynchronous drift, the varying relevance and the label's scarcity. \textbf{The issue of covariate shift} is handled by utilizing the shared parameters between the generative and discriminative phases of denoising autoencoder (DAE). AOMSDA features an open structure in processing multi-source data streams where it copes with \textbf{any concept drifts of source streams}. On the other hand, \textbf{the concept drift of the target domain} is addressed using the node reweighting concept under smoothness assumption. The CMD-based regularizer is put forward to handle \textbf{the issue of varying relevance} where the CMD concept identifies the confidence degree of each source domain. Hence, it mimics the weighting strategy controlling the regularization magnitude as the relevance degree of each source domain. AOMSDA learning strategy works with the assumption of no label of the target domain while only sourcing for labelled samples of source streams. 

The learning policy of AOMSDA is visualized in Fig. \ref{fig_framework}. The training process commences with the estimation of probability density function assumed to follow the Gaussian distribution $p(x)=N(\mu,\sigma^2)$. This process proceeds to the generative phase of the source domains and the target domain minimizing the reconstruction loss. The discriminative phase of the source domain is carried out under shared parameters and minimizes the classification error. The discrminative process makes use of a self-organizing mechanism enabling the node growing and pruning mechanism guided by the probability density function $p(x)$ while the generative phase adopts a fixed structure. The node re-weighting mechanism is carried out afterward. Last but not least, the CMD-based regularizer is carried out to provide the implicit weighting step of each source domain based on the closeness degree of each source stream to the target stream. The simulation protocol of AOMSDA follows \textbf{the prequential test-then-train protocol} where a model is supposed to predict unlabelled samples of the target stream before utilizing them for model updates in an unsupervised fashion. Algorithm \ref{Algorithm_1} offers an overview of AOMSDA's learning policy. 

\subsection{Network Structure of AOMSDA}
AOMSDA is constructed from the autoencoder (AE) \cite{HintonSalakhutdinov2006b,Zhou2012OnlineIF,ASHFAHANI2020297} consisting of two learning phases: generative and discriminative learning phases. The two learning phases are fully coupled where network parameters are shared in the two phases and executed in the lifelong fashion to handle never-ending data streams. The generative phase extracts robust features of original input attributes $X$ and projects it into a low dimensional embedding space $f(.)$ via the encoding mechanism. The decoding mechanism maps the latent features back to the reconstructed space and assures that the original input representation can be reconstructed. The encoding and decoding mechanisms are written as follows:
\begin{equation}
    f_{enc}=s(X_tW_{enc}+b_{enc})
\end{equation}
\begin{equation}
  \hat{X_t}=s(f_{enc}W_{dec}+b_{dec})  
\end{equation}
where $W_{enc}\in\Re^{u\times R},b_{enc}\in\Re^R$ respectively denote the connective weight and bias of the encoder while $W_{dec}\in\Re^{R\times u},b_{dec}\in\Re^u$ respectively label the connective weight and bias of the decoder. $R$ is the number of hidden nodes automatically generated during the structural learning of AOMSDA and $s(.)$ stands for the sigmoid activation function.  

The discriminative phase also known as the classification phase is to map the latent space $f(.)$ to the target space. It is achieved by using the softmax operation as follows: 
\begin{equation}
    \hat{y}=softmax(s(X_tW_{enc}+b_{enc})W_{out}+b_{out})
\end{equation}
where $W_{out}\in\Re^{R\times m},b_{out}\in\Re^{m}$ respectively denote the connective weight and bias of the softmax layer while $softmax(x_i)=\frac{\exp{(x_i)}}{\sum_{o=1}^m\exp{(x_o)}}$. It is seen that the encoder weight and bias, $W_{enc},b_{enc}$, are shared across the discriminative and generative phases. 
\subsection{Parameter Learning Strategy of AOMSDA}
The parameter learning strategy of AOMSDA is devised to resolve the four aforementioned issues of the multistream classification problem under multi-source domains. A joint optimization problem is formulated as follows 
\begin{equation}\label{Lall}
    L_{all}=\sum_{i=1}^{N_s}L_1^i+L_2+\alpha CMD(f_{enc}^{S_i},f_{enc}^T)||W||_2
\end{equation}
where the first loss function $L_1^i$ aims to handle the issue of covariate shift, the second loss function $L_2$ aims to cope with the issue of asynchronous drift and the last term aims to overcome the issue of varying relevance. $\alpha$ is the regularization constant controlling the influence of regularization. All of which are an unconstrained loss function which can be solved directly using the stochastic gradient descent approach. Furthermore, an alternate optimization framework is carried out here where every loss function is minimized alternately. 
\newline\noindent\textbf{Generative and Discriminative Loop:} \textbf{the issue of covariate shift} is handled using the generative and discriminative training phases using shared parameters. This strategy also addresses \textbf{the problem of label's scarcity} since the domain adaptation technique is carried out with the absence of any labelled samples of the target stream. The first loss function of (\ref{Lall}) is formulated as follows:
\begin{equation}
    L_{1}^i=\underbrace{L(x_{S_i},\hat{x}_{S_i})}_{L_{1,1}}+\underbrace{L(y_{S_i}, \hat{y}_{S_i})}_{L_{1,2}}+\underbrace{L(x_T, \hat{x}_T)}_{L_{1,3}}
\end{equation}
where the first term, $L_{1,1}$, the second term, $L_{1,2}$, and the third term $L_{1,3}$ respectively stand for the generative phase of $i-th$ source stream, the discriminative phase of $i-th$ source stream and the generative phase of the target stream. It aims to produce a domain-invariant network handling both the source streams and the target stream equally well. The generative phase of source domains and target domain are driven to minimize the gap between the source and target domains while the discriminative phase of the source domains aims to represent an ideal discriminative representation of the target domain suffering from the absence of any labelled samples. 

The same strategy is also applied in \cite{zhuang2015supervised,ghifary2016deep} where the domain adaptation strategy is formulated as the generative and discriminative training phases of source domain and target domain. The key difference of our approach lies in the extension of this method for streaming context as well as the multi-source domains. Although no direct distance minimization of the two domains exists in the loss function, a domain invariant network is established here because it constructs a feature mapping containing overlapped information between the source and target domains. In other words, the discriminative representation of the source domain where labelled samples exist is retained in the target domain by sharing network parameters in each phase. 

$L(y_{S_i}, \hat{y}_{S_i})$ stands for the discriminative loss function of the source domain. It aims to guarantee a high accuracy of the source domain while preparing for the ideal discriminative representation of the target domain. It can be expressed:
\begin{equation}\label{CE}
    L(y_{S_i}, \hat{y}_{S_i})=-\frac{\sum_{n=1}^{N_i}\sum_{o=1}^m 1(o=y_n^i)log(\hat{y}_n^i)}{N_i}
\end{equation}
where $y_n^i$ denotes the target vector represented as the one-hot vector, $\hat{y}_n^i$ labels the predictive target vector. $1(·)$ return 1 only when the inside logic value is true, $m$ is the number of class labels, and $N_i$ is the size of $i-th$ source stream. (\ref{CE}) is known as the cross entropy loss function.

$L(x_{S_i},\hat{x}_{S_i})$ and $L(x_T, \hat{x}_T)$ refer to the generative loss function of the $i-th$ source domain and the target domain respectively. It minimizes the reconstruction error of the original input representation $x_{S_i},x_T$ from their corrupted version $\tilde{x}_{S_i},\tilde{x}_T$ as per denoising autoencoder (DAE) \cite{vincent2008DAE}. That is, the masking noise is applied here to partially destroy the original input representation $x_{S_i},x_T$ and in turn robust features $f(x_{S_i}),f(x_T)$ can be extracted. In addition to extract robust features, the noise injecting mechanism functions as the regularization mechanism preventing the overfitting problem. $L(x_{S_i,T}\hat{x}_{S_i,T})$ is derived as follows
\begin{equation}
    L(x_{S_i,T},\hat{x}_{S_i,T})=\frac{\sum_{n=1}^{N_{i,T}}(x_{S_i,T}^n-\hat{x}_{S_i,T}^n)^2}{N_{i,T}}
\end{equation}
The MSE loss function is applied here. Using the shared parameters across the generative and discriminative phases allows to resolve the covariate shift problem. That is, previously dissimilar marginal distribution $P_{S_i}(X) \neq P_{S_j}(X) \neq P_{T}(X)$ can be mapped similarly  $P_{S_i}(f(X)) \approx P_{S_j}(f(X)) \approx P_{T}(f(X))$. 
\newline\noindent\textbf{Node Re-weighting Strategy}: the drift handling mechanism of the target stream is difficult due to the absence of any labelled samples. The node re-weighting strategy is integrated here to overcome \textbf{concept drifts of the target stream} where the parameters of encoder is readjusted to arrive at fine-grained feature representation in respect to the concept drift. The node re-weighting mechanism is derived from the smoothness assumption where the predictive outputs should be smooth for similar samples \cite{sun2011two}. That is, it should output similar outputs for adjacent data samples. The second loss function of (\ref{Lall}) is written:
\begin{equation}\label{L2}
    L_2=\sum_{i,j=1,i\neq j}^{N_t}(\hat{y}_i^T-\hat{y}_j^T)^2W_{i,j}
\end{equation}
where $W_{i,j}=exp(-\frac{||x_i-x_j||_2}{2\sigma})$ portrays the similarity of the two samples while $N_t$ is the size of the target stream. (\ref{L2}) can be modified to speed up computation:
\begin{equation}
    L_2=\mathbf{I}^{'}\hat{Y}^{T'}L_t\hat{Y}^T\mathbf{I}
\end{equation}
 where $\mathbf{I}$ is a $c\times 1$ vector only containing 1, $Y^T$ is a $N_t \times m$ matrix in which each row is $\hat{y_i}^T$, and $L_t$ is the graph Laplacian that can be given by $L_t=D-W$. In that equation, $W \in \mathbf{R}^{N_t \times N_t}$ is the similarity matrix which every value denotes the similarity of two samples from the target domain. $D$ is the diagonal matrix, satisfying $D_{i,i}=\sum_{j=1}^nW_{i,j}$. Note that the self-organizing mechanism is not carried out during the generative phase of target stream as in \cite{pratama2019atl} because of the absence of labelled samples. A predictive output does not represent the ground truth which cannot be set as a basis of structural evolution. 
 \newline\noindent\textbf{CMD-based regularization:} this module functions as an implicit weighting strategy of the source domain where the CMD technique \cite{zellinger2017central} finds out the divergence of the $i-th$ source stream and the target stream. The CMD index steers the regularization intensity where low regularization intensity, i.e. accepted, is returned if a source domain is relevant to the target domain whereas high regularization intensity, i.e., rejected, is given to a source domain having low relevance to the target domain. In other words, it mixes source domain information to address the classification problem of the target domain. The CMD technique is performed in the embedding space, \textbf{the transformed space} and is combined with $L_2$ regularization. $CMD(f_{enc}^{S_i},f_{enc}^T)$ is expressed:
 \begin{equation}
    CMD(f_{enc}^{S_i},f_{enc}^T) = \frac{||E(f_{enc}^{S_i})-E(f_{enc}^T)||_2}{|b-a|}+\sum_{k=2}^{K}\frac{||C_k(f_{enc}^{S_i})-C_k(f_{enc}^T)||}{|b-a|^k}
\end{equation}
 where $[a,b]$ is the boundary of activation function, $E(X) = \frac{1}{|X|}\sum_{x\in X}x$, and $C_k(X)=E((x-E(X))^k)$, representing the vector of all $k-th$ order sample central moments. The advantage of CMD as a probabilistic distance measurement of two domains is perceived in the use of high order moments. The dynamic relationship across multi source domains are considered here rather than one source domain. That is, each source stream provides complementary information. It is also worth noting that the CMD measures the relationship of two domains in the embedding space rather than in the original feature space. The CMD regularization strategy is applied to tackle \textbf{the issue of varying relevance}.
\subsection{Structural Learning Strategy of AOMSDA}
AOMSDA features an open structure where its hidden nodes are self-organized from data streams. This procedure takes place in the discriminative phase of source domains where the access of ground truth is available. In other words, this mechanism is used to address \textbf{the concept drift issue in the source stream}. Note that the generalization performance of the target domain is upper bounded by the empirical error of the source domain \cite{ben2010theory}. This mechanism is governed by the network significance (NS) method \cite{ADL,nadine} derived from the bias-variance decomposition. The network significance estimates the network generalization power under a particular probability density function following the normal distribution $NS=\int_{-\infty}^{+\infty}(y_{S_i}-\hat{y}_{S_i})^2p(x)dx;p(x)=N(\mu,\sigma^2)$. The NS formula is derived as follows:
\begin{equation}
\label{NS}
    NS = (E(\hat{y}_{S_i})^2-E(\hat{y}_{S_i}^2))+(E(\hat{y}_{S_i})-y_{S_i})^2=Var+Bias^2
\end{equation}
The key in deriving the bias and variance expression lies in $E[\hat{y}_{S_i}]$ which comes from $E[\hat{y}_{S_i}]=softmax(\int_{-\infty}^{+\infty}s(X_{S_i}W_{enc}+b_{enc})p(x)dxW_{out}+b_{out})$. It is seen that it depends on the integral operation over sigmoid function which does not have an exact integral solution. The sigmoid function is approachable using the probit function and the integral of probit function is another probit function \cite{murphy2012probabilistic}. This aspect leads to the final expression of $E[\hat{y}_{S_i}]$ as follows
\begin{equation}
    E[\hat{y}_{S_i}]=softmax(s(\frac{W_{enc}\mu}{\sqrt{1+\pi\sigma^2/8}}+b_{enc})W_{out}+b_{out})
\end{equation}
where $\mu,\sigma$ are the mean and standard deviation of the normal distribution $N(\mu,\sigma^2)$ which can be recursively calculated. The same strategy is applicable to the variance expression where under the i.i.d condtion $E[\hat{y}_{S_i}^2]=E[\hat{y}_{S_i}].E[\hat{y}_{S_i}]$.

The hidden node growing and pruning conditions are signalled by the statistical process control (SPC) approach \cite{GamaDataStream} adapting to the concept drifts of the source domain and used frequently in the context of anomaly detection. A node is added in the case of high bias indicating the underfitting situation while the pruning condition is triggered by a high variance condition signifying the overfitting condition. The node growing and pruning conditions are formulated:
\begin{equation}
    \label{growing}
    \mu_{bias}^t + \sigma^t_{bias}\geq\mu_{bias}^{min}+\pi\sigma^{min}_{bias}\rightarrow Growing
\end{equation}
\begin{equation}
    \label{pruning}
    \mu_{var}^t + \sigma^t_{var}\geq\mu_{var}^{min}+2\chi\sigma^{min}_{var}\rightarrow Pruning
\end{equation}
where the confidence level of SPC is controlled by $\pi=1.3exp(-(Bias(\hat{y}))^2)+0.7$ and $\chi=1.3exp(-(Var(\hat{y})))+0.7$. This strategy is meant to assure that the growing and pruning mechanisms are carried out when the bias and the variance are high. In particular, the confidence degree of the node growing condition revolves around $68.3\%$ to $95.2\%$ while the pruning condition is in the range of $95.2\%$ to $99.99\%$. $\mu_{min}^{bias},\sigma_{min}^{bias}$ are reset if (\ref{growing}) is satisfied. On the other hand, $\mu_{min}^{bias},\sigma_{min}^{bias}$ are reset if (\ref{pruning}) is observed. The term $2$ is inserted in (\ref{pruning}) to avoid the direct-pruning-after adding situation impeding the structural evolution. Furthermore, (\ref{growing}) sets a high bias condition leading to the introduction of a new node. A new node is created and initialized using the Xavier's initialization strategy. The high variance condition is alarmed by (\ref{pruning}) where the node pruning process is applied to the least-contributing node having the lowest statistical contribution \cite{ADL}. Although the area of evolving intelligent system (EIS) for data streams has been an active research topic for many years as surveyed in \cite{EFSreview}, to the best of our knowledge, the issue of multistream classification problem remains an uncharted territory. 

\begin{table*}[t!]
    \centering
    \caption{Properties of the datasets}
    \begin{centering}
    \scalebox{0.8}{\begin{tabular}{c|c|c|c|c|c|c}
    \hline
    \hline
          Dataset &F &C &$SS$  &TS &CP(\%) &Char 
          \tabularnewline
    \hline
          Weather &3 &2 &4.5K &4.5K &[31.38, 68.62] &Non-Stationary
         \tabularnewline
    \hline
          Sea &3 &2 &25K &25K  &[63.04, 36.96] &Non-Stationary
          \tabularnewline
    \hline
          Hyperplane &4 &2 &30K &30K &[50.06, 49.94] &Non-Stationary
          \tabularnewline
    \hline
          KDDCup &41 &2 &125K  &125K &[80.09, 19.91] &Non-Stationary
          \tabularnewline
    \hline
          Kitti &55 &8 &6.25K  &6.25K &\makecell[c]{[74.57, 8.30,\\ 1.75, 7.77,\\ 0.93, 3.85, \\1.63, 1.19]} &Non-Stationary
          \tabularnewline
    \hline
          Susy &18 &2 &1.25M &1.25M &[54.24, 45.76] &Stationary
          \tabularnewline
    \hline
         Hepmass &27 &2 &2.6M  &2.6M &[50.00, 50.00] & Stationary
         \tabularnewline
    \hline
        OQC &48 &3 &5.75K &5.75K &[37.5, 26.11, 36.38] &Non-Stationary\tabularnewline
    \hline
    \hline
    \end{tabular}}
    \par\end{centering}
    \centering{F: The dimensions of features; C: The number of classes; $SS$: The number of samples from one of the three source domains, all three source domains contain the same number of samples; $TS$: The number of samples from the target domain; CP: Class proportion; Char: Characteristics}
    \label{G2}
\end{table*}

\begin{table}[t]
      \caption{Hyperparameters of consolidated algorithms}
    \centering
    \begin{tabular}{c|c|c|c|c}
    \hline
    \hline
    &\multirow{1}{*}{{Melanie}} &\multirow{1}{*}{{ATL}} &\multirow{1}{*}{{MDAN}} & \multirow{1}{*}{{AOMSDA}}\tabularnewline
    \hline
         Learning Rate $\zeta$ &Nan &0.01 &0.01 &0.01 \tabularnewline
         Network Structure &Nan &Nan &[1000 500 100] &Nan \tabularnewline
         Tradeoff Parameters $\alpha$ &Nan &Nan &Nan &1.0\tabularnewline
         Time forgetting factor $\theta$ &0.9 &Nan &Nan &Nan\tabularnewline
         Performance index $\lambda$ &0.05 &Nan &Nan &Nan\tabularnewline
         Gamma $\gamma$ &Nan &Nan &10 &Nan\tabularnewline
         \hline
         \hline
    \end{tabular}
    \label{hyperparameters}
\end{table}
\begin{table*}[t]
\caption{Numerical Results of Consolidated Algorithm}
\begin{centering}
\label{G3} \scalebox{0.6}{ 
\begin{tabular}{|c|c|c|c|c|c|c|c|c|}
\hline
  &  & \multirow{1}{*}{{Melanie}} & \multirow{1}{*}{{ATL}} & \multirow{1}{*}{{$MDAN_s 15$}} &\multirow{1}{*}{{$MDAN_s 1$}} &\multirow{1}{*}{{$MDAN_h 15$}}&\multirow{1}{*}{{$MDAN_h 1$}}&\multirow{1}{*}{{AOMSDA}}\tabularnewline
\hline 
\parbox[t]{25mm}{\multirow{2}{*}{\rotatebox[origin=c]{0}{Weather}}}
&CR    &\textbf{77.74}  &$74.26$ &$75.32$   &$63.95$ &$74.47$ &$75.44$ &$76.55$ \tabularnewline
&Trt   &\textbf{2.81}  &$41.08$ &$7.09$  &$1.27$ &$19.53$  &$1.28$ &$34.69$ \tabularnewline
&HN   &\textbf{Nan}  &$129.00$ &$1000,500,100$ &$1000,500,100$ &$1000,500,100$  &$1000,500,100$ &$18.20$ \tabularnewline
\hline 
\parbox[t]{25mm}{\multirow{2}{*}{\rotatebox[origin=c]{0}{Sea}}}
&CR    &$89.18$  &$88.59$ &$88.93$ &$71.10$ &$87.32$ &$81.24$  &\textbf{90.23} \tabularnewline
&Trt   &$5.89$  &$161.57$ &$112.41$ &$7.29$ &$113.40$ &$7.73$   &\textbf{195.77} \tabularnewline
&HN   &$Nan$  &$54.67$ &$1000,500,100$ &$1000,500,100$ &$1000,500,100$  &$1000,500,100$   &\textbf{19.60} \tabularnewline
\hline 
\parbox[t]{25mm}{\multirow{2}{*}{\rotatebox[origin=c]{0}{Hyperplane}}}
&CR    &$86.38$  &$86.01$ &$89.11$ &$84.36$ &\textbf{89.21} &$87.61$   &$88.03$\tabularnewline
&Trt   &$7.40$  &$133.59$  &$128.73$ &$8.97$ &\textbf{204.63} &$9.01$ &$236.18$  \tabularnewline
&HN   &$Nan$ &$26.67$ &$1000,500,100$ &$1000,500,100$ &\textbf{1000,500,100}  &$1000,500,100$ &$24.30$   \tabularnewline
\hline 
\parbox[t]{25mm}{\multirow{2}{*}{\rotatebox[origin=c]{0}{KDDCup}}}
&CR    &$95.24$ &$99.49$ &$97.27$ &$94.73$ &$97.32$ &$92.19$  &\textbf{99.76} \tabularnewline
&Trt   &$43.13$& $8446.46$ &$574.36$ &$36.41$ &$550.36$ & $36.56$&\textbf{1176.99}    \tabularnewline
&HN   &$Nan$  &$195.44$ &$1000,500,100$ &$1000,500,100$ &$1000,500,100$  &$1000,500,100$  &\textbf{32.00}  \tabularnewline
\hline 
\parbox[t]{25mm}{\multirow{2}{*}{\rotatebox[origin=c]{0}{Kitti}}}
&CR    &$50.29$  &$52.88$ &\textbf{74.89*} &$5.28$ &\textbf{74.89*} &$8.40$ &$67.79$  \tabularnewline
&Trt   &$24.15$  &$309.25$ &\textbf{27.88} &$1.77$ &\textbf{26.65} &$1.87$  &$54.89$  \tabularnewline
&HN   &$Nan$  &$1599.50$ &\textbf{1000,500,100} &$1000,500,100$ &\textbf{1000,500,100}  &$1000,500,100$   &$33.30$ \tabularnewline
\hline 
\parbox[t]{25mm}{\multirow{2}{*}{\rotatebox[origin=c]{0}{Susy}}}
&CR    &$72.94$  &$62.48$ &$59.85$ &$58.62$ &$62.72$ &$60.24$ &\textbf{79.41*}  \tabularnewline
&Trt   &$420.00$  &$138512.21$ &$15000.43$ &$375.93$ &$9058.22$ &$367.47$  &\textbf{10094.66}  \tabularnewline
&HN   &$Nan$  &$745.00$&$1000,500,100$ &$1000,500,100$ &$1000,500,100$  &$1000,500,100$   &\textbf{14.30}  \tabularnewline
\hline 
\parbox[t]{25mm}{\multirow{2}{*}{\rotatebox[origin=c]{0}{Hepmass}}}
&CR    &$72.01$  &$74.18$ &$71.34$ &$83.06$ &$71.29$ &$83.74$  &\textbf{86.21*} \tabularnewline
&Trt   &$6074.00$  &$346407.74$ &$50452.89$ &$3097.022$ &$13840.07$ &$798.69$   &\textbf{21816.93} \tabularnewline
&HN   &$Nan$  &$2968.00$&$1000,500,100$ &$1000,500,100$ &$1000,500,100$  &$1000,500,100$   &\textbf{16.80} \tabularnewline
\hline  
\parbox[t]{25mm}{\multirow{2}{*}{\rotatebox[origin=c]{0}{OQC}}}
&CR    &$33.33$  &$49.47$ &$36.92$ &$33.49$ &$35.44$ &$31.14$  &\textbf{68.56*}\tabularnewline
&Trt   &$3.21$  &$1060.62$ &$21.67$ &$1.62$ &$21.52$ &$1.35$   &\textbf{28.75} \tabularnewline
&HN   &$Nan$  &$5497.56$ &$1000,500,100$ &$1000,500,100$ &$1000,500,100$  &$1000,500,100$   &\textbf{32.20} \tabularnewline
\hline  
\end{tabular}} 
\par\end{centering}
\centering{CR: Classification accuracy rate; Trt: Training time; HN: The number of hidden nodes; $MDAN_s 15$: MDAN soft version with 15 epochs; $MDAN_h 15$: MDAN hard version with 15 epochs; $MDAN_s 1$: MDAN soft version wiht 1 epoch; $MDAN_h 1$: MDAN hard version with 1 epoch};*: Statistically significant compared with other algorithms.
\end{table*} 

\begin{table}[t]
    \centering
    \caption{Recall and Precision of AOMSDA}\scalebox{0.9}{
    \begin{tabular}{|c|c|c|}
    \hline
         &  Recall &Precision  \tabularnewline
         \hline
   Weather &0.49 & 0.24\tabularnewline
   \hline
   Sea &0.91 & 0.91\tabularnewline
   \hline
   Hyperplane &0.89 &0.85 \tabularnewline
   \hline
    KDDCup &0.99  &0.99 \tabularnewline
    \hline
     Kitti &\makecell[c]{[0.80, 0.07, 0.15, 0.48,\\ 0.01, 0.003, 0.11, 0]} &\makecell[c]{[0.78, 0.11, 0.19, 0.37,\\ 0.08, 0.005, 0.04, 0]} \tabularnewline
     \hline
      Susy &0.61 &0.82 \tabularnewline
      \hline
    Hepmass    &0.74 &0.76 \tabularnewline
   \hline
   OQC    &[0.50, 0.66, 0.81] &[0.66, 0.57, 0.69] \tabularnewline
   \hline
    \end{tabular}}
    \label{recall_pecision}
\end{table}

\begin{table*}[!t]
\caption{Ablation study}
\begin{centering}
\label{G4} \scalebox{0.9}{ 
\begin{tabular}{|c|c|c|c|c|c|c|}
\hline
  &  & \multirow{1}{*}{{Original}} & \multirow{1}{*}{{A}} & \multirow{1}{*}{{B}} &\multirow{1}{*}{{C}} &\multirow{1}{*}{{D}} \tabularnewline
\hline 
\parbox[t]{18mm}{\multirow{3}{*}{\rotatebox[origin=c]{0}{Weather}}}
&CR    &\textbf{76.55}  &$75.27$ &$75.52$   &$75.19$  & $74.44$\tabularnewline
&Trt   &\textbf{34.69}  &$41.08$ &$28.07$  &$34.25$ &$38.58$\tabularnewline
&HN   &\textbf{18.20}  &$15.20$ &$10$ &$16.80$  &$20.22$\tabularnewline
\hline 
\parbox[t]{18mm}{\multirow{3}{*}{\rotatebox[origin=c]{0}{Sea}}}
&CR    &\textbf{90.23}  &$90.15$ &$89.98$ &$90.08$ & $89.37$\tabularnewline
&Trt   &\textbf{195.77}  &$176.36$ &$152.73$ &$189.04$ &$188.65$ \tabularnewline
&HN   &\textbf{19.60}  &$18.40$ &$10$ &$24.70$  &$126.53$\tabularnewline
\hline 
\parbox[t]{18mm}{\multirow{3}{*}{\rotatebox[origin=c]{0}{Hyperplane}}}
&CR    &$88.03$  &$88.14$ &$86.91$ &\textbf{88.15} &$86.69$\tabularnewline
&Trt   &$235.18$  &$224.26$ &$183.91$ &\textbf{223.92} &$239.04$\tabularnewline
&HN   &$24.30$ &$25.30$  &$10$ &\textbf{23.50}  & $100.11$ \tabularnewline
\hline 
\parbox[t]{18mm}{\multirow{3}{*}{\rotatebox[origin=c]{0}{KDDCup}}}
&CR    &$99.76$ &\textbf{99.77} &$99.75$ &$99.69$ &$99.76$ \tabularnewline
&Trt   &$1176.99$ &\textbf{968.36} &$895.96$ &$1047.07$ &$5112.98$\tabularnewline
&HN   &$32.60$  &$33.40$ &$10$ &$28.90$ &$37.84$\tabularnewline
\hline 
\parbox[t]{18mm}{\multirow{3}{*}{\rotatebox[origin=c]{0}{Kitti}}}
&CR    &$67.79$  &\textbf{68.69} &$67.84$ &$66.48$ & $62.61$\tabularnewline
&Trt   &$54.89$  &\textbf{47.74} &$46.32$ &$54.54$ &$230.33$\tabularnewline
&HN   &$33.30$  &\textbf{31.20} &$10$ &$34.60$ &$378.48$\tabularnewline
\hline 
\parbox[t]{18mm}{\multirow{3}{*}{\rotatebox[origin=c]{0}{Susy}}}
&CR    &$79.41$  &$79.42$ &$80.40$ &$78.58$ &\textbf{81.98}\tabularnewline
&Trt   &$10094.66$  &$9156.43$ &$8737.89$ &$10463.14$  &\textbf{59795.05}\tabularnewline
&HN   &$14.30$  &12.80 &$10$ &$23.40$ &\textbf{397.62}\tabularnewline
\hline 
\parbox[t]{18mm}{\multirow{3}{*}{\rotatebox[origin=c]{0}{Hepmass}}}
&CR    &$86.21$  &$86.14$ &$86.30$ &\textbf{87.22}  &$85.65$\tabularnewline
&Trt   &$21816.93$  &$19851.13$ &$13452.13$ &\textbf{21996.33} &$253451.03$\tabularnewline
&HN   &$16.80$  &$16.90$ &$10$ &\textbf{18.90} &$497.34$ \tabularnewline
\hline
\parbox[t]{18mm}{\multirow{3}{*}{\rotatebox[origin=c]{0}{OQC}}}
&CR    &$68.56$  &$67.23$ &$56.53$ &$69.36$  & \textbf{69.37}\tabularnewline
&Trt   &$29.75$  &$19.48$ &$20.04$ &$21.96$ & \textbf{256.13} \tabularnewline
&HN   &$32.20$  &$34.91$ &$10$ &$35.46$ & \textbf{749.64}\tabularnewline
\hline
\parbox[t]{18mm}{\multirow{1}{*}{\rotatebox[origin=c]{0}{Mean}}}
&CR &\textbf{82.07} &$81.58$ &$80.40$ &$81.84$ &$81.23$\tabularnewline
\hline  
\end{tabular}} 
\par\end{centering}
\centering{CR: Classification accuracy rate; Trt: Training time; HN: The number of hidden nodes; Original: Original experiments; A: Experiments without Node re-weighting strategy; B: Experiments without structural learning strategy; C: Experiments without CMD regularization; D: Replace Gaussian distribution with AGMM.}
\end{table*}

\begin{table*}[t]
\caption{The influence of the number of source domains}
\begin{centering}
\label{G5} \scalebox{1}{ 
\begin{tabular}{|c|c|c|c|c|c|}
\hline
  &  & \multirow{1}{*}{{1 source}} & \multirow{1}{*}{{3 sources}} & \multirow{1}{*}{{5 sources}} &\multirow{1}{*}{{7 sources}} \tabularnewline
\hline 
\parbox[t]{25mm}{\multirow{3}{*}{\rotatebox[origin=c]{0}{Weather}}}
&CR    &$72.21$  &$76.55$ &$75.89$   &$79.73$  \tabularnewline
&HN   &$8.20$  &$18.25$ &$23.60$ &$27.20$  \tabularnewline
\hline 
\parbox[t]{25mm}{\multirow{3}{*}{\rotatebox[origin=c]{0}{Sea}}}
&CR    &$88.17$  &$90.23$ &$89.32$ &$88.49$  \tabularnewline
&HN   &$17.40$  &$19.60$ &$26.40$ &$33.60$  \tabularnewline
\hline 
\parbox[t]{25mm}{\multirow{3}{*}{\rotatebox[origin=c]{0}{Hyperplane}}}
&CR    &$86.30$  &$88.03$ &$87.73$ &$87.75$ \tabularnewline
&HN   &$6.00$ &$24.30$  &$37.40$ &$50.40$    \tabularnewline
\hline 
\parbox[t]{25mm}{\multirow{3}{*}{\rotatebox[origin=c]{0}{KDDCup}}}
&CR    &$99.76$ &$99.76$ &$99.69$ &$99.69$ \tabularnewline
&HN   &$14.60$  &$32.60$ &$45.20$ &$66.40$ \tabularnewline
\hline 
\parbox[t]{25mm}{\multirow{3}{*}{\rotatebox[origin=c]{0}{Kitti}}}
&CR    &$53.59$  &$67.79$ &$68.31$ &$58.76$ \tabularnewline
&HN   &$27.80$  &$33.30$ &$45.00$ &$46.80$ \tabularnewline
\hline 
\parbox[t]{25mm}{\multirow{3}{*}{\rotatebox[origin=c]{0}{Susy}}}
&CR    &$75.70$  &$79.41$ &$81.21$ &$84.50$\tabularnewline
&HN   &$9.40$  &$14.30$  &$24.20$ &$49.20$ \tabularnewline
\hline 
\parbox[t]{25mm}{\multirow{3}{*}{\rotatebox[origin=c]{0}{Hepmass}}}
&CR    &$80.83$  &$86.21$ &$88.45$ &$89.44$  \tabularnewline
&HN   &$13.60$  &$16.80$ &$31.00$ &$49.80$  \tabularnewline
\hline  
\parbox[t]{25mm}{\multirow{3}{*}{\rotatebox[origin=c]{0}{OQC}}}
&CR    &$67.35$  &$68.56$ &$77.05$ &$79.36$  \tabularnewline
&HN   &$32.58$  &$32.20$ &$50.14$ &$57.37$  \tabularnewline

\hline  
\end{tabular}} 
\par\end{centering}
\centering{CR: Classification accuracy rate; HN: The number of hidden nodes}
\end{table*}

\section{Numerical Study}
This section outlines our numerical validation where AOMSDA is tested in eight problems and compared against state-of-the art algorithms. The efficacy of AOMSDA is examined using an ablation study demonstrating the advantage of each learning component of AOMSDA. In addition, the effect of the number of source streams is also studied where AOMSDA's learning performance is evaluated under varying numbers of source streams.   
\subsection{Simulation Protocol} Our numerical study is simulated in respect to the prequential test-then-train protocol where a model is supposed to predict a data stream first before exploiting it for model updates. That is, the whole dataset is partitioned into a number of data batches where an initial model is crafted from the first data batch to address the cold start problem. A data batch is split in respect to the Gaussian distribution where 3 source streams and 1 target stream are arranged \cite{MSC}. Note that AOMSDA is general for any number of source streams and the effect of source streams is also studied in this paper. The Gaussian distribution is referred here to induce \textbf{the issue of covariate shift}. That is, the probability of each sample is calculated as $\exp{\frac{(x-\mu)^2}{2\sigma^2}}$ where data samples are sorted in a descending order. The first source stream is built upon the top $N_i$ samples followed by the second source stream, the third source stream and so on up to the $N_S$ source streams while the target stream is arranged as the remainder of data samples. For simplicity, every stream is set to be an equal size $N_1=N_2=N_i=...=N_S=N_T$. Only source streams contain labelled samples while the target stream is left unlabelled - \textbf{scarcity of labelled samples.} The numerical evaluation is performed independently per a data batch to reflect the performance under drifts where the overall numerical results are averaged across all data batches. 

\subsection{Datasets}
Eight datasets, namely, Weather \cite{DitzlerImbalanced}, Sea \cite{SEA}, Hyperplane \cite{Bifet07learningfrom}, KDDCUP \cite{KDDCup}, SUSY \cite{Baldi2014SearchingFE}, Forest Cover\cite{forestcover}, Kitti \cite{Geiger2012CVPR} and Online Quality Classification (OQC) \cite{NADINE++}, are used to numerically validate the advantage of our algorithm. All datasets except susy and hepmass contain concept drifts leading to \textbf{the issue of asynchronous drift} and \textbf{the issue of varying relevance}. The properties of datasets are summarized in Table \ref{G2}.  The characteristics of eight datasets are elaborated as follows:
\newline\noindent\textbf{Weather:} This dataset describes one-step-ahead prediction whether or not rain occurs. It records the weather data over 50 years containing the yearly seasonal change as well as the long-term climate change. 

\noindent\textbf{SEA:} this problem is a synthetic dataset having both recurring and abrupt concept drift. The binary classification satisfies $f_1+f_2\geq \theta$, where $\theta$ changes three times as $\theta=4\rightarrow 7\rightarrow 4 \rightarrow 7$, which causes concept drift.

\noindent\textbf{Hyperplane:} this problem is designed to predict the position of $d-th$ dimensional hyperplane satisfying $\sum_{j=1}^d\omega_jx_j>\omega_0$. The transition period when the second distribution replaces the first one causes the gradual drift in this dataset.

\noindent\textbf{Kitti:} this dataset constitutes a real-world computer vision problem derived from the autonomous driving problem in the city of Kalsruhe, Germany. The underlying goal is to perform 3D object detection where the object is captured by two high-resolution color and grayscale video cameras.  

\noindent\textbf{KDDCUP:} this dataset describes a network intrusion detection problem where the simulation of various network attacks results in the non-stationary characteristic. 

\noindent\textbf{SUSY:} it is a popular big dataset containing 5 million instances, which presents a signal process causing super-symmetric particles. Although it is stationary, this dataset has a big characteristic thereby being able to test the algorithm's characteristic in overcoming a large-scale problem.  

\noindent\textbf{Hepmass:} this dataset describes the separation of particle-producing collisions from a background with a big size that contains more than 10 million samples. As with the susy dataset, this dataset is deployed to examine the algorithm's performance in the large-scale setting.


\noindent\textbf{Online Quality Classification (OQC)}: this problem features a quality detection problem of a transparent part of scent-emitting USB device manufactured by an injection-molding machine \cite{NADINE++}. There are two common defects of the transparent part: short-forming and weaving rendering it a three-class classification problem. Moreover, this dataset is non-stationary in nature because the injection speed and the holding pressure are varied during the production phase. The quality prediction problem is guided by 48 input attributes recording different machine parameters.

\subsection{Baseline}
AOMSDA is compared against three algorithms: Melaine \cite{du2019multi}, ATL \cite{pratama2019atl} and MDAN \cite{zhao2018multiple,zhao2018adversarial} where their detailed characteristics are discussed:
\newline\noindent\textbf{Melanie} is a multistream classification method under multi-source domains. It is developed from the ensemble concept and features the drift handling aptitude. 
\newline\noindent\textbf{ATL} is akin to Melaine where it is designed to solve the multistream classification problem but under a single source domain environment. Because it is not designed for the multi-source domains, ATL regards multi-source streams as a single source stream while the evaluation phase is drawn from its performance in the target stream. Compared to AOMSDA, ATL is not equipped by the CMD regularizer and the node reweighting strategy. 
\newline\noindent\textbf{MDAN} is an offline domain adaptation technique for multi-source domains setting. This work adopts the idea of adversarial domain adaptation and is seen as a state-of-the art algorithm for multi-source domains adaptation. MDAN is set into four configurations here: hard-max for 15 epochs, hard-max for 1 epoch, soft-max for 15 epochs, soft-max for 1 epoch. Note that MDAN is an offline algorithm having significant advantage compared to AOMSDA because of its iterative nature.

All algorithms are simulated under the same computational environments by running their published codes which can be easily adapted to the same simulation environments. Direct comparison with reported results are not possible to be done because of different simulation and computational environments. To the best of our knowledge, the multi-stream classification problem under multi-source domain is also an uncharted territory in the existing literature making direct comparison with the reported results of the three algorithms difficult to be done.

\subsection{Hyper-parameters}
 The learning rates of ATL, MDAN and AOMSDA  are set the same to ensure fair comparisons $\zeta=0.01$ while the tradeoff parameter of AOMSDA is set as $\alpha=1.0$. MDAN's structure is configured as a three hidden layer network with 1000 nodes of the first layer, 500 nodes of the second layer and 100 nodes of the third layer as per their default setting in their codes while $\gamma$ of MDAN is fixed at $10$. On the other hand, the time forgetting factor $\theta$ and the performance index $\lambda$ of MELANIE are respectively selected at 0.9 and 0.05. The hyperparameters of all algorithms are displayed in the Table \ref{hyperparameters}. These hyper-parameters are fixed throughout all our numerical studies to ensure fair comparison and to demonstrate that AOMSDA is non-adhoc. Note that AOMSDA and ATL run fully in the one-pass training scenario with a single epoch. 

\subsection{Numerical Results}
Numerical results of benchmarked algorithms are reported in Table \ref{G3} where the algorithm's performance is evaluated in three facets: classification rates (CR), training time (Trt) and the number of hidden nodes (HN). The classification rate of an algorithm is measured by an average accuracy of all data batches, while the training time is obtained from the total training time during the training process and the hidden node is taken from the final number of hidden nodes. All algorithms are simulated under the same computational environments and executed five times. The numerical results in Table \ref{G3} are reported as an average across five runs. Since the benchmarked algorithms are developed from different languages: MDAN (Python), Melanie (JAVA), ATL and AOMSDA (Matlab), AOMSDA's execution time is directly comparable to only ATL. 

AOMSDA outperforms other algorithms in five of eight problems: Sea, KDDCup, Susy, Hepmass and OQC. AOMSDA beats other algorithms with noticeable margins in two big data problems: Hepmass ($3\%$) and Susy ($5\%$). Our method is inferior to Melanie by around 1\% in the Weather problem, but it is well-known that the performance of neural-network based algorithm compromises in this problem due to the uncertainty issue of the Weather problem as also depicted in the ATL and MDAN numerical results. AOMSDA is also inferior to MDAN with 15 epochs in the Hyperplane and Kitti problems by 1\% and 7\% difference. This result should be interpreted carefully because MDAN is run through many epochs, 15 epochs due to an offline algorithm. MDAN performance is compromised if it undergoes a one-pass training as with AOMSDA. AOMSDA also outperforms other algorithms with significant margin ($10\%$) in the OQC problem. Our numerical results are confirmed with a statistical test, the t-test, where AOMSDA produces statistically superior results in the Susy, Hepmass and OQC problems ($P<0.05$). Furthermore, it is observed that Melanie does not deliver statistically significant results compared to AOMSDA in the Weather problem.

Table \ref{recall_pecision} reports the precision and recall of AOMSDA across 8 datasets. It is observed that the gap between precision and recall are not high except for the weather problem where three exists about $20\%$ gap. This problem is known to be challenging for NN-based algorithms. Low gap between precision and recall means unbiased prediction toward one of the classes. Another interesting finding is in the KDDCup problem where AOMSDA delivers the same precision and recall although this problem has skewed class proportion. Nevertheless, we acknowledge that the class imbalance issue still requires in-depth study and is beyond the scope of this paper.  

Since ATL does not have any mechanism to mix the multi-source streams, the performance of ATL drops significantly. In the context of execution time, AOMSDA demonstrates an improved performance compared to ATL. Note that the runtime of AOMSDA is only comparable to ATL because both are developed under MATLAB environments. Slow computation time of ATL is mostly caused by the use of AGMM in calculating the network bias and variance. This approach incurs expensive computational burden and is often unstable in the case of high input dimension. 

Although Melanie is developed to solve the multi-source streams, it is only capable of surpassing other algorithms in the Weather problem well-known to be difficult to handle by NN-based algorithms. The numerical results of MDAN is not directly comparable to AOMSDA because it gains a full access of the whole dataset before process runs as well as it performs iterative training across many epochs. Albeit these advantages, MDAN outperforms AOMSDA in only two cases: Kitti and Hyperplane where these results are achieved using 15 epochs. AOMSDA consistently beats MDAN with a single epoch in all cases.   

\begin{figure}
    \centering
    \includegraphics[width=3in]{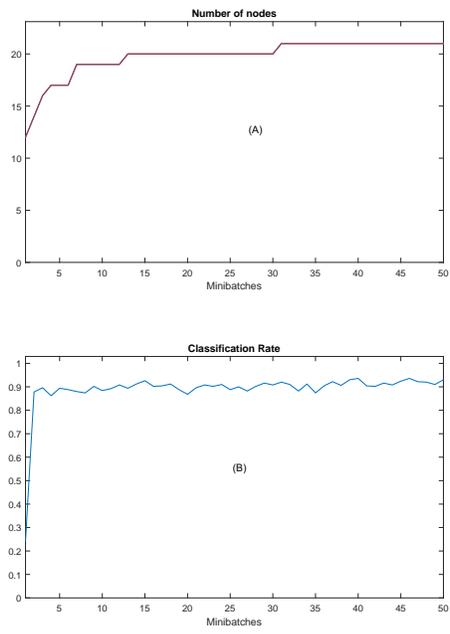}
    \caption{(A) trace of hidden nodes in the SEA problem; (B) trace of classification rates in the SEA problem}
    \label{sea}
\end{figure}

Fig. \ref{sea} presents the trace of hidden node and classification rate of AOMSDA in the SEA problem. It is seen that AOMSDA is capable of dynamically inserting its hidden nodes from data streams addressing concept drifts of the source streams. Note that concept drifts might occur in both source and target streams in different time points. On the other hand, it is also seen that the dynamic of classification rate is relatively stable. This finding implies that concept drifts do not undermine model's generalization. Performance losses due to the concept drifts can be recovered quickly. 

\subsection{Ablation Study}
This section studies the influence of each learning component on the final performance of AOMSDA. AOMSDA is configured into four versions: (A) the absence of the node re-weighting mechanism; (B) the absence of structural learning mechanism; (C) the absence of CMD-based regularizer; (D) the integration of AGMM \cite{pratama2019atl} into AOMSDA. The ablation study is carried out with all datasets to arrive at solid conclusions. The numerical results are presented in Table \ref{G4} where the average of classification rates across all problems is put forward at the bottom of the Table. 

It is evident from Table \ref{G4} that the current configuration of AOMSDA delivers the best-performing results on average. The absence of node re-weighting strategy decreases the numerical results in Weather, Sea, Hepmass and OQC problems. That is, the node re-weighting approach is capable of refining the numerical results. On the other hand, the absence of structural learning mechanism deteriorates the numerical results in the Weather, Sea, KDDCup, Susy and OQC problems significantly. This finding confirms the efficacy of structural learning in handling the concept drifts. AOMSDA's performance is compromised with significant margins if the CMD-based regularizer is deactivated. Its performance drops in the Weather, Sea, KDDCup, Kitti and Susy problems. This fact portrays the importance of CMD-based regularizer in weighting the multi-source streams. The use of AGMM in model D does not improve the performance of AOMSDA. The accuracy worsens in most of the cases while the complexity significantly mounts in all of the cases. This is caused by the fact that multiple hidden nodes are directly added if the node growing condition is triggered. The use of AGMM also imposes expensive computational complexity as depicted in the training time of Model D. Note that the deactivation of a learning module only leads to minor performance's increases in some cases. On the contrary, it results in substantial performance losses in other cases.
\subsection{The number of Source Streams}
The effect of the number of source streams is analyzed here. It answers two fundamental questions whether multi-source streams improves the learning performance of the single-source stream and AOMSDA is general for any number of source streams.  AOMSDA's learning performance is examined using 1, 3, 5 and 7 source streams respectively where our numerical study is undertaken with all datasets to arrive at valid conclusion. Numerical results are reported in the Table \ref{G5}.   

The advantage of multi-source domains is obvious in Table \ref{G5}. That is, the accuracy of AOMSDA under multi-source streams setting improves from its single-source stream version in all problems. On the other hand, the learning performance of AOMSDA is stable across any number of source streams. It is perceived that the classification accuracy of AOMSDA consistently increases as the increase of the number of source streams in weather, SUSY, Hepmass and OQC. Different numbers of source streams do not change the accuracy significantly in other four problems but remains better than the accuracy of single source stream configuration. In realm of complexity, the increase of the number of source streams causes model's structural complexity to grow. This issue emerges as a result of the structural learning strategy of AOMSDA taking place in the discriminative phase of the source domain.       

\section{Conclusion}
This paper offers a solution of multi-stream classification problem under multi-source domains with algorithmic development of automatic online multi-source domain adaptation (AOMSDA). AOMSDA combines the domain adaptation technique and the drift handling mechanism while featuring the mixing strategy of multi-source streams. The domain adaptation strategy relies on a generative and discriminative loop of DAE discovering an overlapped region of the source streams and the target stream thus addressing the covariate shift problem. The idea of CMD-based regularization is integrated to cope with the varying relevance of source domains to the target domain. It functions as some sort of weighting mechanism to every source domain where it controls the regularization intensity when learning a source stream. The asynchronous drift problem is overcome by the node re-weighting strategy under the smoothness assumption. That is, a model should output similar prediction for adjacent samples. Last but not least, AOMSDA features a self-organizing structure where the hidden nodes are dynamically grown and pruned from data streams when learning source domain in the discriminative fashion. Our numerical study demonstrates that AOMSDA performs favourably compared to the state-of-the art algorithms in five of eight study cases. It is also confirmed with the ablation study where the current configuration of AOMSDA delivers better performance than other four configurations on average. Furthermore, the advantage of multi-source streams is depicted where it delivers an improved performance compared to the single source stream configuration and AOMSDA is general for any number of source streams. Our future works will be devoted to study the problem of cross-domain transfer learning.     
\section{Acknowledgement}
This work is supported by Ministry of Education, Republic of Singapore, Tier 1 Grant. 


.

\bibliography{mybibfile}

\end{document}